\documentclass[journal,twocolumn]{IEEEtran}
\usepackage{epsfig,makeidx,color, subfigure, multirow}
\usepackage{amsmath,amssymb,bbm}
\usepackage[ruled,linesnumbered]{algorithm2e}
\usepackage{algpseudocode}
\usepackage{cite,graphicx,lipsum}
\usepackage{enumerate}
\usepackage{amsfonts}
\usepackage{booktabs}
\usepackage{siunitx}
\usepackage{empheq}
\usepackage[table,xcdraw]{xcolor}
\usepackage[switch,pagewise]{lineno}
\usepackage{hyperref}
\hypersetup{
        colorlinks = true,
        citecolor=blue,
}
\usepackage{graphicx}
\usepackage{booktabs}
\usepackage{graphicx}
\usepackage[normalem]{ulem}
\usepackage{tikz}
\usepackage{xcolor}

\DeclareRobustCommand\encircle[1]{\tikz[baseline=(char.base)]{\node[shape=circle,fill,inner sep=1pt] (char) {\textcolor{white}{#1}}}}

\def\l{\left}
\def\r{\right}
\def\({\l(}
\def\){\r)}
\def\[{\l[]}
\def\]{\r]}

\def\vx{{\mathbf x}}

\def\vz{{\mathbf z}}

\def\vc{{\mathbf c}}

\def\vtheta{\boldsymbol \theta}
\def\vphi{\boldsymbol \phi}
\def\vgamma{\boldsymbol \gamma}

\begin{document}


\title{\fontsize{22}{28}\selectfont Semantics Alignment via Split Learning for \\Resilient Multi-User Semantic Communication}

\author{Jinhyuk Choi, $^\dagger$Jihong Park, $^\ddagger$Seung-Woo Ko, $^\dagger$Jinho Choi, $^*$Mehdi Bennis, and Seong-Lyun Kim\\
\thanks{J. Choi and S.-L. Kim are with the School of Electrical and Electronic Engineering, Yonsei University, Seoul 03722, Korea.} 
\thanks{$^\dagger$J. Park and J. Choi are with the School of Information Technology, Deakin University, Geelong, VIC 3220, Australia.}
\thanks{$^\ddagger$S.-W. Ko is with the Department of Smart Mobility Engineering, Inha University, Incheon 21999, Korea.}
\thanks{$^*$M. Bennis is with the Centre for Wireless Communications, University of Oulu, Oulu 90014, Finland.}
\thanks{J. Park, S.-W. Ko, and S.-L. Kim are corresponding authors (email: jihong.park@deakin.edu.au, swko@inha.ac.kr, slkim@yonsei.ac.kr).}
}

\maketitle

\begin{abstract}
Recent studies on semantic communication commonly rely on neural network (NN) based transceivers such as deep joint source and channel coding (DeepJSCC). Unlike traditional transceivers, these neural transceivers are trainable using actual source data and channels, enabling them to extract and communicate semantics. On the flip side, each neural transceiver is inherently biased towards specific source data and channels, making different transceivers difficult to understand intended semantics, particularly upon their initial encounter. To align semantics over multiple neural transceivers, we propose a distributed learning based solution, which leverages split learning (SL) and partial NN fine-tuning techniques. In this method, referred to as SL with layer freezing (SLF), each encoder downloads a misaligned decoder, and locally fine-tunes a fraction of these encoder-decoder NN layers. By adjusting this fraction, SLF controls computing and communication costs. Simulation results confirm the effectiveness of SLF in aligning semantics under different source data and channel dissimilarities, in terms of classification accuracy, reconstruction errors, and recovery time for comprehending intended semantics from misalignment.

\end{abstract}

\ifCLASSOPTIONonecolumn
\baselineskip 26pt
\fi

\section{Introduction}

\subsection{Semantic Communication using Neural Transceivers}
While recent advances in machine learning have transformed communication system's design principles \cite{pieee21park},
it can be argued that semantic communication (SC) is an area that significantly benefits from the application of machine learning techniques
\cite{bourtsoulatze2019deep,qin2021semantic,seo23}. While classical communication focuses on transferring the bit representations of data over noisy channels to reconstruct the original data \cite{ShannonWeaver:1949}, SC aims to convey meaningful or \emph{semantic representations (SRs)} of the data, tailored for specific tasks such as classification, control, and other tasks \cite{bourtsoulatze2019deep,qin2021semantic,seo23}. To enable SC built upon classical communication operations, one promising approach is via artificial intelligence (AI) native transceiver designs that utilize a neural network (NN) as a trainable end-to-end transceiver, as elaborated next.

An NN is ideally a universal function approximator \cite{hornik1989multilayer}, and has a great potential in simultaneously emulating multiple functionalities that are tantamount to a composite function. Following this principle, it is possible to train an NN to emulate (i) source coding and (ii) channel coding functionalities in classical communication systems \cite{Hoydis:Asilomar18}. In addition to (i) and (ii), an NN can simultaneously emulate two new SC functionalities: (iii) pre-processing for extracting semantics and (iv) post-processing to solve a downstream task. Deep joint source and channel coding (DeepJSCC) is one promising approach that can concurrently emulate (i)-(iv) by using the autoencoder (AE) NN architecture \cite{bourtsoulatze2019deep,qin2021semantic} consisting of a set of encoder layers (\textsf{ENC}) and its paired decoder layers (\textsf{DEC}). The AE NN of DeepJSCC is trained for a given task, after which SRs are generated from raw data at an \textsf{ENC} and delivered to the \textsf{DEC} producing outputs for the given task.
\vspace{-15pt}

\begin{figure}\centering
\includegraphics[width=\columnwidth]{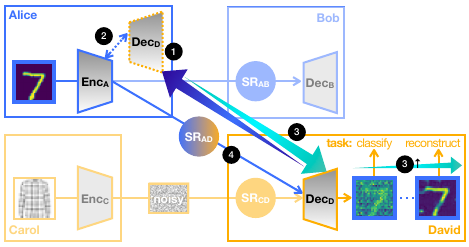}
\caption{A schematic illustration of split learning with layer freezing (SLF) with two DeepJSCC AEs: Alice-Bob AE and Carol-David AE trained under heterogeneous datasets (MNIST and Fashion-MNIST) and different levels of channel noise.} \label{Fig1}
\vspace{-10pt}
\end{figure}

\subsection{Misalignment Multi-User Semantic Communication}
While effective in various tasks ranging from image reconstruction \cite{bourtsoulatze2019deep} to visual question answering \cite{xie2021taskVQA}, due to its NN architecture, one fundamental limitation of DeepJSCC is its inherent bias towards \emph{a)} source (training) data and \emph{b)} \textsf{ENC}-\textsf{DEC} channel characteristics during training. To illustrated this by an example, as shown in Fig. 1, consider two DeepJSCC AEs, namely the AE $\textsf{ENC}_1$-$\textsf{DEC}_1$ of Alice and Bob, and the other AE $\textsf{ENC}_2$-$\textsf{DEC}_2$ of Carol and David, which were trained under different data source and/or channel environments. After training, Alice and Bob communicate their intended semantics, and so do Carol and Bob. However, between Alice and David (or equivalently Carol and Bob), the SRs generated by Alice may not always be interpreted as intended at David
due to the absence of joint training for the cross-pair, specifically $\textsf{ENC}_1$-$\textsf{DEC}_2$.
This \emph{semantics misalignment problem} is particularly critical in mobile scenarios whereby any newly encountered transceivers are unlikely to be interoperable, restricting the scalability of multi-user SC.
\vspace{-15pt}

\subsection{Aligning Semantics via Split Learning with Layer Freezing}
In this article, we focus on the aforementioned semantics misalignment problem for multi-user SC, and aim to make SC robust against dissimilar source data and/or channels with low latency as well as low communication and computation costs. We tackle this problem by aligning the semantics between different DeepJSCC transceivers, inspired from split learning (SL) that trains multiple NNs while shuffling the split-segments of the NNs \cite{Vepakomma:2018:Splita}. To this end, we propose a novel DeepJSCC fine-tuning method, coined \emph{SL with layer freezing (SLF)}. As depicted in Fig.~\ref{Fig1}, with two DeepJSCC transceivers, the operations of SLF are summarized into the following steps. 

\begin{enumerate}
    \item[\encircle{\footnotesize 1}] \textbf{Decoder Downloading}: Alice first downloads David's decoder $\textsf{DEC}_2$ through \emph{background communication}, which is in contrast to exchanging SRs through \emph{foreground communication}.
    
    \item[\encircle{\footnotesize 2}] \textbf{Local Fine-Tuning}: By connecting the downloaded decoder with its local encoder,  Alice locally re-trains $\textsf{ENC}_1$-$\textsf{DEC}_2$ using its own source data and the Alice-David channel statistics that can be obtained during background communication under channel reciprocity.
    
    \item[\encircle{\footnotesize 3}] \textbf{Fine-Tuned Decoder Uploading}: After obtaining the fine-tuned pair $\textsf{ENC}_1'$-$\textsf{DEC}_2'$, Alice uploads the re-trained $\textsf{DEC}_2'$ back to David.

    \item[\encircle{\footnotesize 4}] \textbf{Aligned SR Transmission}: Finally, Alice transmits SRs generated from $\textsf{ENC}_1'$, and David can decode it using its $\textsf{DEC}_2'$. 
\end{enumerate}

During the fine-tuning process in \encircle{\footnotesize 2}, Alice can partially fix the layers of $\textsf{DEC}_2$, and re-train only the remainder. Fine-tuning computation cost commonly increases with the number of trainable layers. Moreover, only the fine-tuned layers need to be uploaded from Alice to David in the background communication. Therefore, while the $\textsf{DEC}_2$ downloading latency remains the same in \encircle{\footnotesize 1}, the number of frozen layers decreases the fine-tuning computation latency in \encircle{\footnotesize 2} and the uploading latency in \encircle{\footnotesize 3}, resulting in longer \emph{recovery time}, defined as the end-to-end SR alignment latency during \encircle{\footnotesize 1} -- \encircle{\footnotesize 3}. 

On the other hand, for an image reconstruction task, our experiments show that the number of frozen layers increases the reconstruction errors after \encircle{\footnotesize 4}. Consequently, there is a trade-off between reconstruction errors and recovery time, which can be balanced for a given task. For instance, classification tasks may not require high-fidelity reconstruction, allowing Alice to freeze more layers.
\vspace{-10pt}

\subsection{Contributions}

The major contributions of this work are summarized as follows.
\begin{itemize}
    \item We propose SLF, an NN fine-tuning technique for aligning the semantics between two DeepJSCC transceivers trained under dissimilar source data and/or channels.

    \item Focusing on the impact of the number of frozen layers, we delve into recovery time and goals for two different tasks, i.e., mean squared error (MSE) for reconstruction and accuracy for classification, under different levels of source and channel dissimilarities.
    
    \item By simulations, we corroborate that SLF works successfully under various semantic misalignment scenarios. Furthermore, the trade-off between recovery time and task-specific operation in the simulations further emphasizes the importance of our proposed SLF.
\end{itemize}
\vspace{-7pt}

\begin{figure*}[t!]
    \centering
        \subfigure{
        \includegraphics[width=.98\columnwidth]{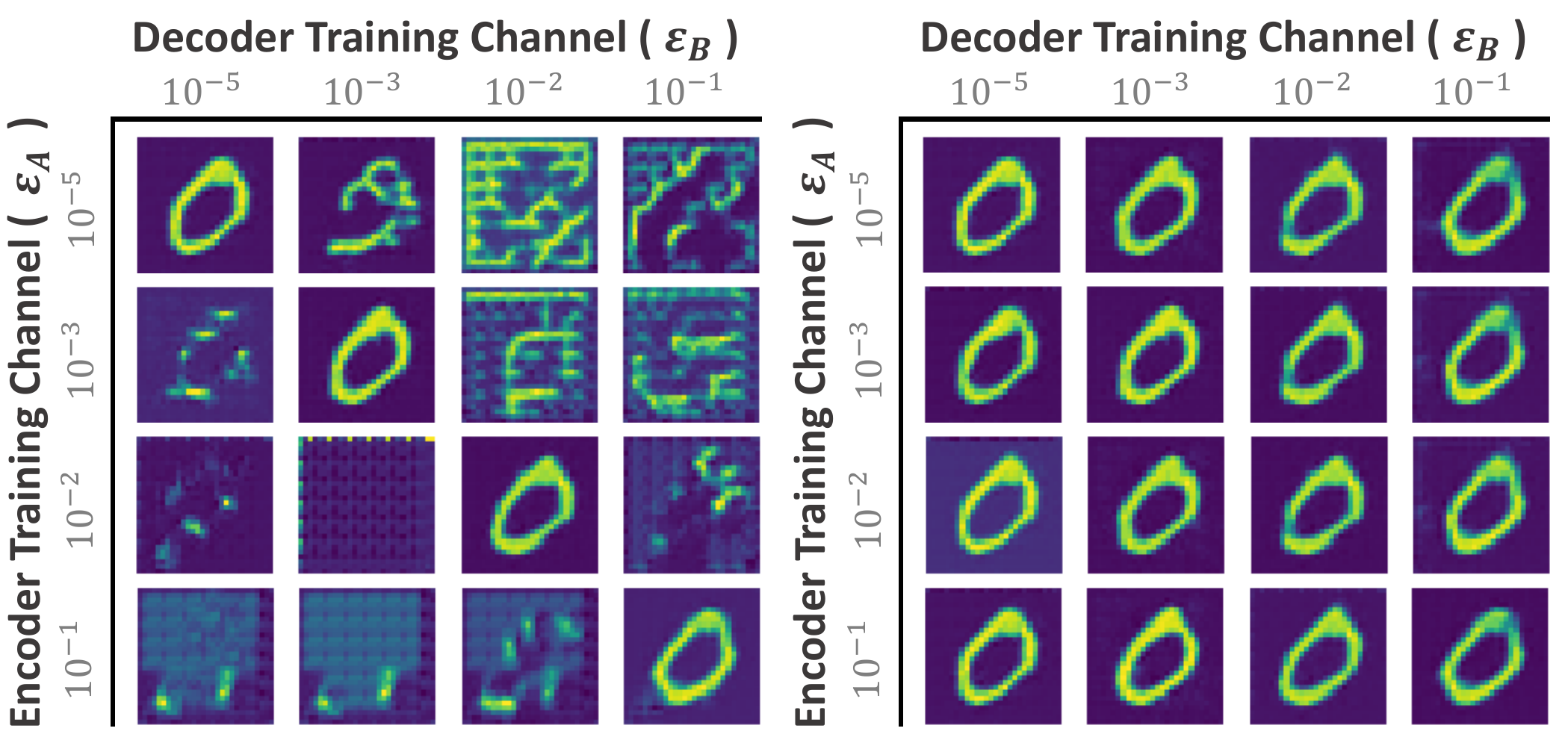}
        }
        \subfigure{
        \includegraphics[width=.98\columnwidth]{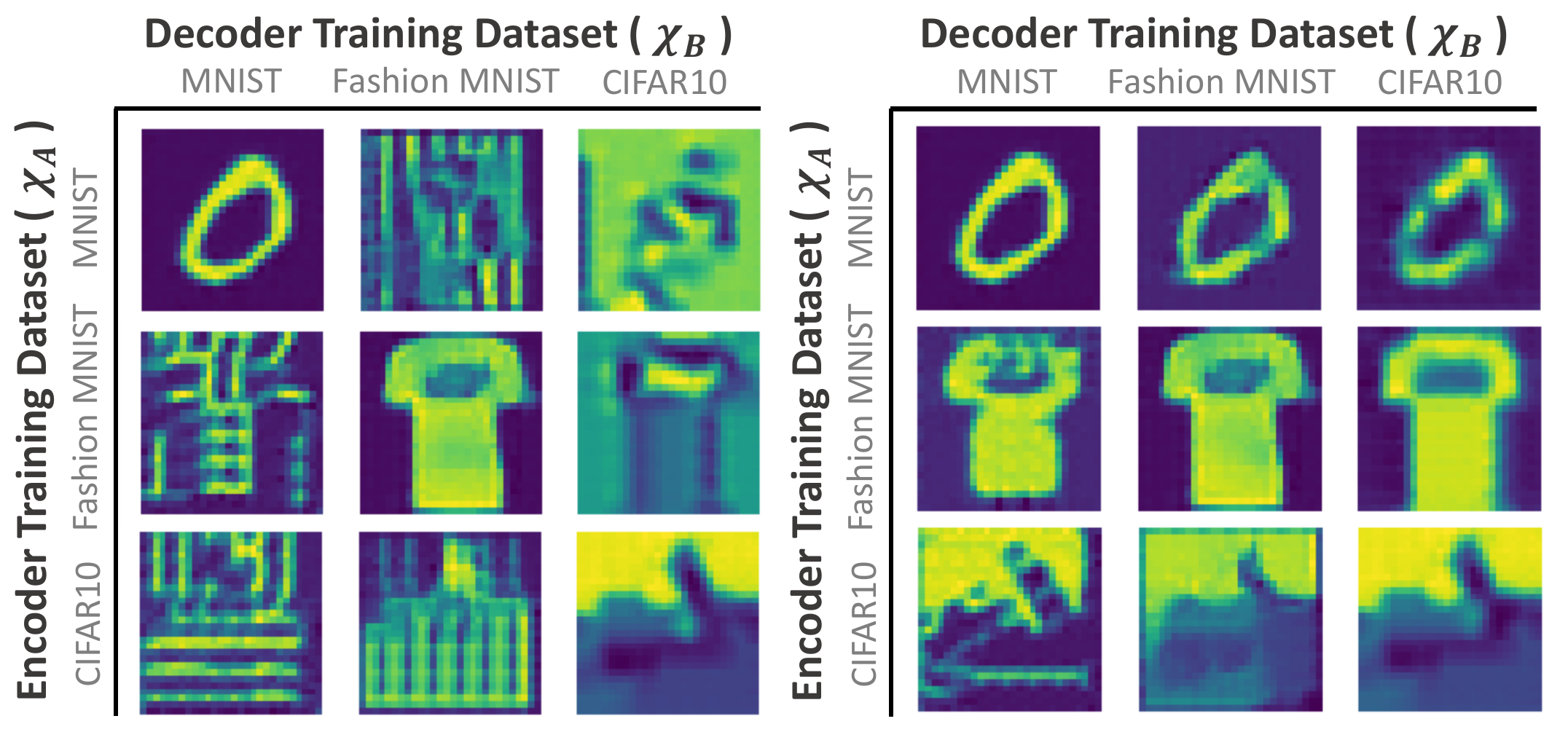}
        }
    \caption{Reconstructed images under different encoder-decoder training channels (1st and 2nd), and under different encoder-decoder training source data (3rd and 4th). Without SLF (1st and 3rd), off-diagonal images visualize the impact of semantics misalignment, which is fixed by SLF (2nd and 4th).
    }  \label{CrossSC Problem example}
    \vspace{-15pt}
\end{figure*}

\section{System Descriptions} \label{Sec:System}
The network under study comprises different pairs of DeepJSCC transceivers $\textsf{TRX}_i$ and $\textsf{TRX}_j$ that are identically constructed on a vector-quantized variational AE (VQ-VAE) architecture \cite{van2017neural} for a common task, while each transceiver is independently pre-trained under different source data and channel characteristics. 
VQ-VAE is a well-established NN architecture, and is capable of performing joint source and channel coding \cite{tung2022deepjscc}. To make VQ-VAE generate task-specific SRs while improving architectural reusability for different tasks, we consider a tripartite VQ-VAE by adding a set of layers into the standard bipartite VQ-VAE for encoding and decoding, as detailed next. 

\textbf{Transceiver Structure}.\quad
Each transceiver $\textsf{TRX}_i=[\vtheta_i, \vphi_i, \vgamma_i]$ with $i\in\{1,2\}$ is an NN that sequentially processes the following three different functions: $\textsf{ENC}_i(\cdot)$ for source-channel encoding, $\textsf{DEC}_i(\cdot)$ for source-channel decoding, and $\textsf{Task}(\cdot)$ for task-specific operations, which are parameterized by three blocks of NN weights $\vtheta_i$, $\vphi_i$, and $\vgamma_i$, respectively. Each $\textsf{TRX}_i$ includes a transmitter $\textsf{TX}_i$ and its paired receiver~$\textsf{RX}_i$. 

\textbf{SR Encoding}.\quad
$\textsf{TX}_i$ stores $\vtheta_i$ and source data samples $\vx_i$ in a local dataset $\mathcal{X}_i$. The encoding of $\vx_i$ through NN layers is described as a function $\textsf{ENC}_{ij}(\cdot)$ mapping $\vx_i$ into the SR $\vz_{ij}$ that is transmitted to $\textsf{RX}_j$, i.e.,
\begin{align}
\vz_{ij} = \textsf{ENC}_{ij}(\vx_i),
\end{align}
where the first and second subscripts of $\vz_{ij}$ identify $\textsf{TX}_i$ and $\textsf{RX}_j$, respectively. Note that $\textsf{ENC}_{ij}(\cdot)$ depends not only on $\textsf{TX}_i$ but also on  $\textsf{RX}_j$, since its encoder-decoder is concurrently trained. Following the standard VQ-VAE, within $\textsf{ENC}_{ij}(\cdot)$, $\vx_i$ is first mapped into a latent variable ${\vz}_{ij}'$, followed by vector quantizing ${\vz}_{ij}'$ into an $\vz_{ij}=\vc_{k^\star} \in \mathcal{C}_{ij}$, where $k^\star := \arg\min_{\vc_k\in\mathcal{C}_{ij}} ||  {\vz}_{ij}' - \vc_k ||_2$ with a trainable codebook $\mathcal{C}_{ij}=\{\vc_1, \vc_2, \cdots, \vc_K \} $. Consequently, the transmitted SR $\vz_{ij}$ is composed of the elements in $K$ codewords.

\textbf{Channel Model}.
The transmitted SR ${\vz}_{ij}$ is distorted by a noisy channel between  $\textsf{TX}_i$ and $\textsf{RX}_j$, which is modeled using the $K$-ary discrete memoryless channel (DMC) \cite{cover1999elements}. For a given DMC crossover probability $\varepsilon_{ij}$, the received SR $\hat{\vz}_{ij}$ is determined by the following transition probability:
\begin{align}
\Pr(\hat{\vz}_{ij}= \vc_{k}| \vz_{ij} = \vc_{k^\star})= 
\begin{array}{ll}
\left\{ \begin{array}{ll}
1-\varepsilon_{ij}, & \text{if $k=k^\star$} \cr 
\frac{\varepsilon_{ij}}{K-1}, & \text{otherwise}. \end{array}\right.
\end{array}
\end{align}
In other words, the received SR $\hat{\vz}_{ij}$ is identical to the transmitted SR ${\vz}_{ij}=\vc_{k^\star}$ with probability $1-\varepsilon_{ij}$, and otherwise becomes one of other $K-1$ codewords with equal probability. Here, the channel statistics can be characterized by $\varepsilon_{ij}$ that increases with outage probability or equivalently decreases with the signal-to-noise ratio, as elaborated in \cite{nemati2022all}.

\textbf{SR Decoding}.\quad
At  $\textsf{RX}_j$, it receives the distorted SR $\hat{\vz}_{ij}$, and yields the reconstructed sample $\hat{\vx}_{ij}$ using the decoding function $\textsf{DEC}_{ij}(\cdot)$ as follows:
\begin{align}
\hat{\vx}_{ij}=\textsf{DEC}_{ij}(\hat{\vz}_{ij}) .
\end{align}
As a result, the end-to-end commmunication is summarized as $\vx_i \overset{\textsf{ENC}_{ij}}{\to} \vz_{ij} \overset{\text{DMC}}{\to} \hat{\vz}_{ij} \overset{\textsf{DEC}_{ij}}{\to} \hat{\vx}_{ij}$. The decoded sample $\hat{\vx}_{ij}$ at $\textsf{RX}_j$ can be different from the original sample $\vx_i$ at $\textsf{TX}_i$ due not only to the channel noise but also to $\textsf{DEC}_{ij}$ that was not jointly trained with $\textsf{ENC}_{ij'}$ if $i\neq j'$. The latter warrants the need for addressing the semantics misalignment problem.

\begin{figure*}[t!]
    \centering
        \subfigure[Reconstruction task.]{
        \includegraphics[width=.49\columnwidth]{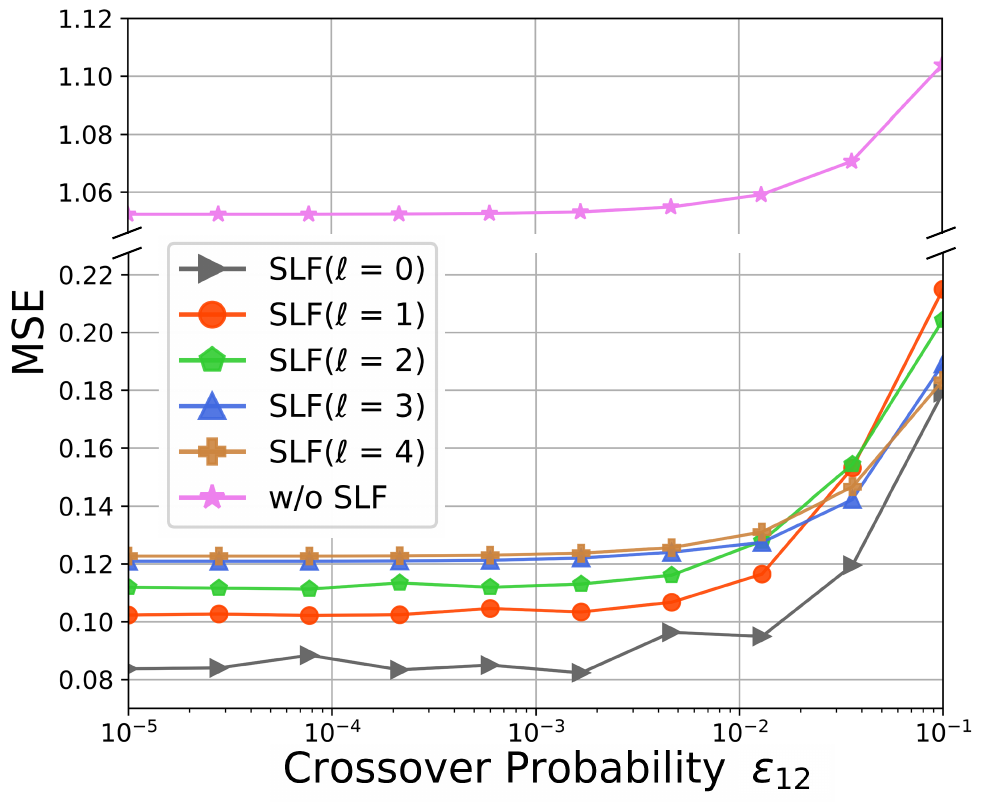}
        \includegraphics[width=.49\columnwidth]{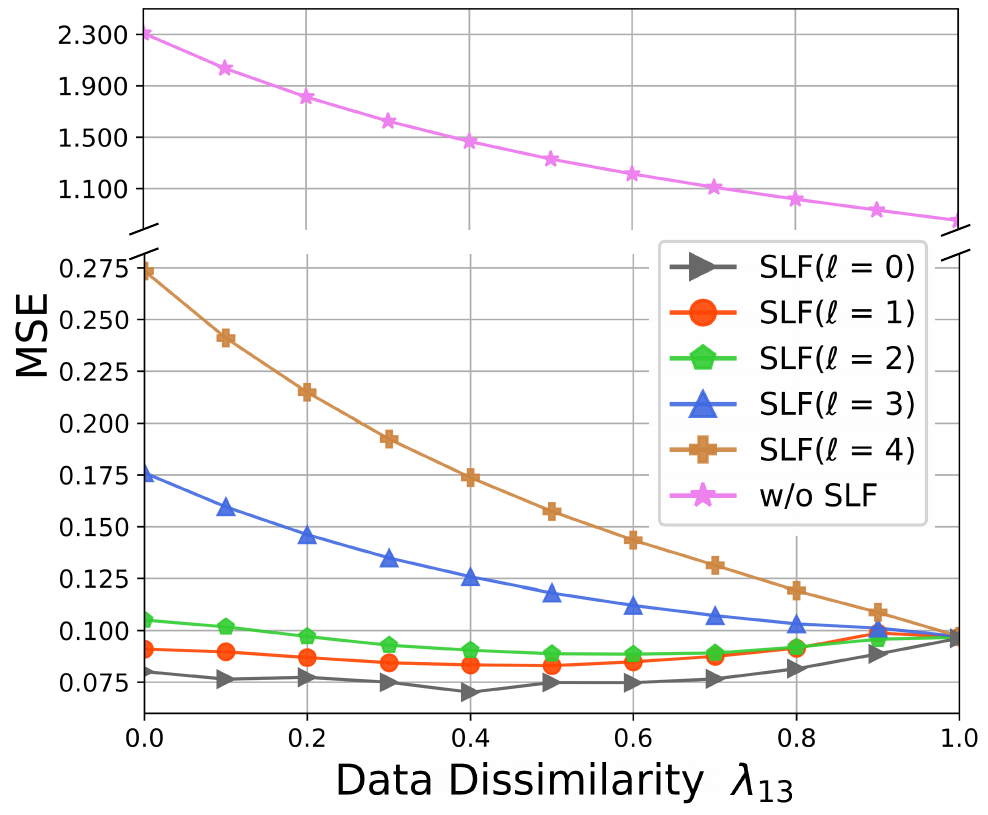}
        \label{SLF graph with CSC(channel characteristic)}
        }
        \subfigure[Classification task.]{
        \includegraphics[width=.49\columnwidth]{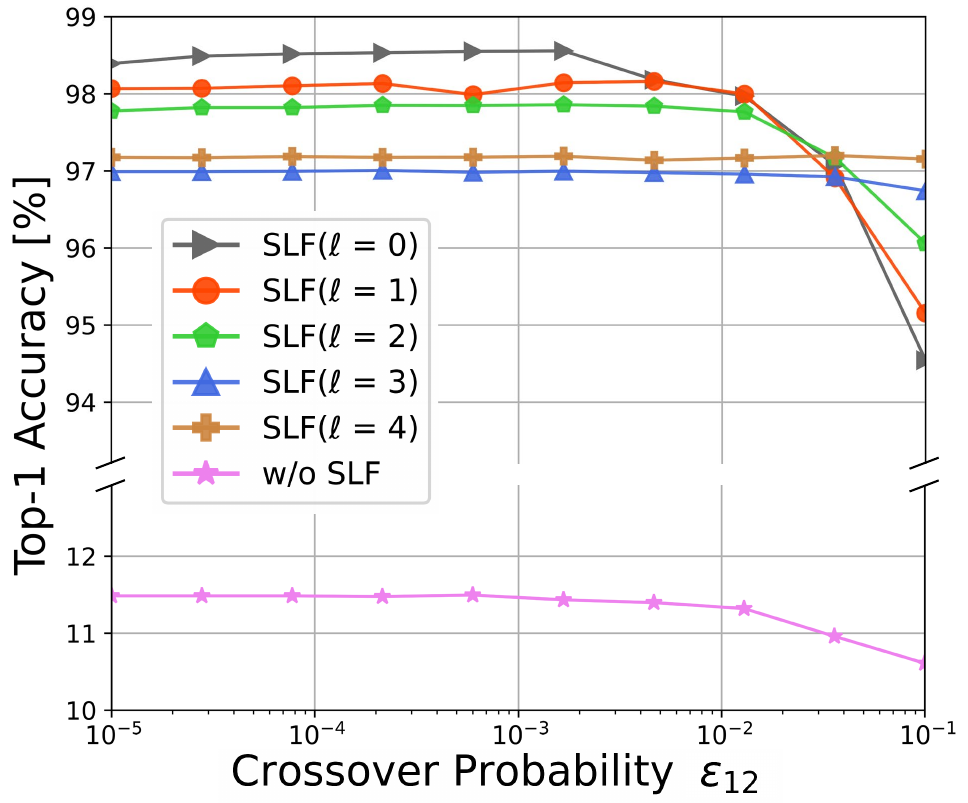}
        \includegraphics[width=.48\columnwidth]{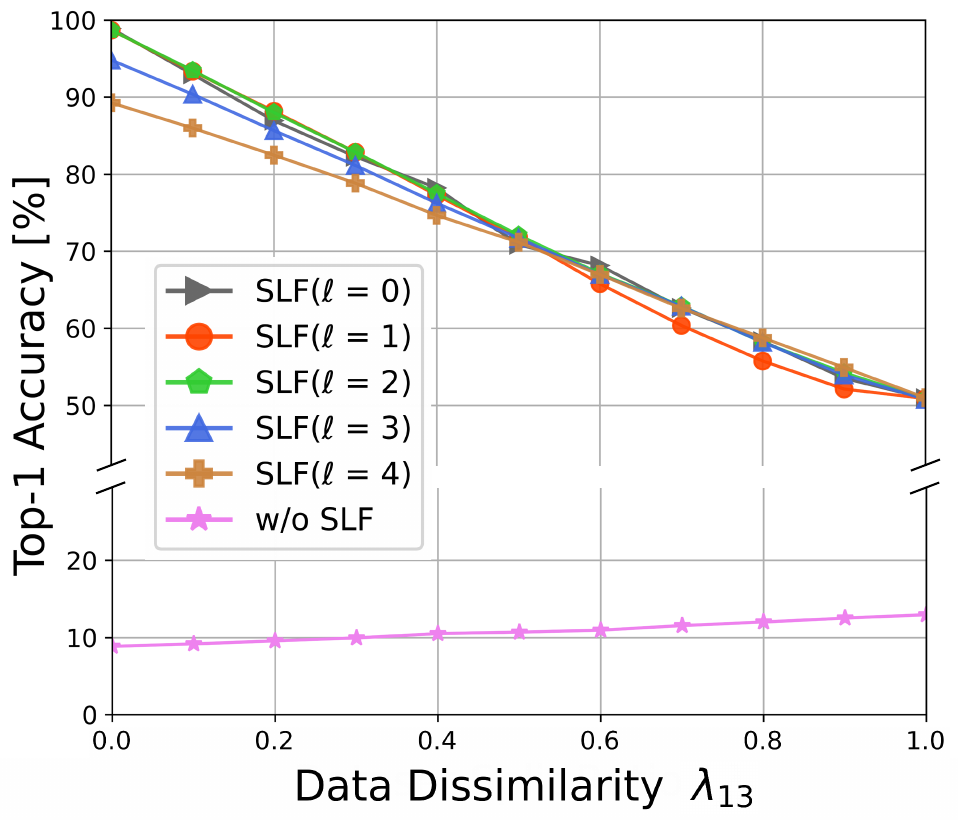}
        \label{SLF graph with CSC(channel characteristic) with classification task}
        }
    \vspace{-10pt}
    \caption{
    Task-specific performance of SLF, i.e., MSE for reconstruction and top-1 accuracy for classification, under different DMC channel cross-over probability $\varepsilon_{12}$ and source data dissimilarity $\lambda_{13}$.} \label{Graph 3}
    \vspace{-15pt}
\end{figure*}

\textbf{Task Effectiveness}.
The $\textsf{RX}_j$ stores $\vgamma_j$ that utilizes $\hat{\vx}_{ij}$ to carry out a task-specific decision-making $\textsf{Task}(\hat{\vx}_{ij})$. We consider two different tasks, source data sample reconstruction and the sample's label classification. 
For the reconstruction task, $\textsf{TRX}_{ij}$ aims to minimize the MSE, $\mathbb{E}\left[||\vx_i - \hat{\vx}_{ij}||^2\right]$, between the source sample $\vx_i$ and the decoded sample $\hat{\vx}_{ij}$. In this case, we have $\vgamma_j=\emptyset$ and the structure of $\textsf{TRX}_{ij}$ boils down to the standard bipartite VQ-VAE, i.e., $\textsf{TRX}_{ij}=[\vtheta_i,\vphi_j]$, and $\textsf{Task}(\hat{\vx}_{ij})=\textsf{DEC}_{ij}(\hat{\vz}_{ij})=\hat{\vx}_{ij}$. On the other hand, for the classification task, we have $\vgamma_j \neq \emptyset$, and $\textsf{TRX}_{ij}$ aims to maximize the top-1 accuracy of the predicted label $\hat{y}_{ij}=\textsf{Task}(\hat{\vx}_{ij})$, for the given ground-truth label $y_i$ associated with $\vx_i$.
\vspace{-10pt}

\section{Split Learning with Layer Freezing}
\subsection{Motivation -- Challenges in Semantics Alignment}
Suppose that there are two independently pre-trained DeepJSCC transceivers $\textsf{TRX}_i$ and $\textsf{TRX}_j$ with $i\neq j$. The $\textsf{TX}_i$ of $\textsf{TRX}_i$ intends to communicate with $\textsf{RX}_j$ of $\textsf{TRX}_j$. This misaligned SC is unlikely to be successful, in that $\textsf{TX}_i$'s $\vtheta_i$ and its original codebook $\mathcal{C}_{ii}$, as well as $\textsf{RX}_j$'s $\vphi_j$ and $\vgamma_j$ are biased towards their separate pre-trained environments, in terms of its source data (i.e., $\mathcal{X}_i$ and $\mathcal{X}_j$) as well as channel characteristics (i.e., $\varepsilon_{ii}$ and $\varepsilon_{jj}$). Indeed, the off-diagonal examples in Figs.~\ref{CrossSC Problem example} show that misaligned SC fails in both the reconstruction task and classification task due to dissimilar channels and source data, respectively.

To make $\textsf{TX}_i$ and $\textsf{RX}_j$ interoperable, a na\"ive solution is to re-train a new transceiver $\textsf{TRX}_{ij}=[\vtheta_i, \vphi_j, \vgamma_j]$ through communication between $\textsf{TX}_i$ and $\textsf{RX}_j$. However, this incurs non-negligible additional communication cost until re-training convergence. Furthermore, it may also violate data privacy, as it should share the local dataset $\mathcal{X}_i$ of $\textsf{TX}_i$ with $\textsf{RX}_j$ at which the training loss is calculated by comparing $\textsf{RX}_j$'s output with $\mathcal{X}_i$; for instance, comparison with the original source sample $\vx_i$ for reconstruction or the ground-truth label $y_i$ for classification. 

Meanwhile, noticing that similar communication cost and data privacy issues have recently been tackled in the domain of distributed learning \cite{pieee21park}, one may attempt to tackle this semantics misalignment problem using federated learning (FL) \cite{mcmahan2017communication}, wherein clients train their local models in collaboration by averaging model parameters, without exchanging their private training datasets. Unfortunately, the semantics misalignment problem coincides with an extreme case of imbalanced data distributions across clients, also known as the non-independent and identically distributed (non-IID) data problem, under which the effectiveness of FL is significantly compromised \cite{zhao2018federated}. In fact, our preliminary study in \cite{park2022federated} demonstrates that FL only marginally improves convergence speed in re-training without any gain in accuracy, although it incurs significant communication cost due to exchanging model parameters per re-training iteration.
\vspace{-15pt}

\subsection{SLF for Aligning Semantics in Multi-User SC}
Alternatively, to align semantics in multi-user SC, we propose a novel fine-tuning method, termed SLF. SLF leverages SL \cite{vepakomma2018split} to divide each transceiver into its encoder and decoder segments, followed by exchanging and fine-tuning different combinations of these segments. As visualized in Fig.~\ref{Fig1}, SLF operates in the following four steps.

\begin{enumerate}
\item[\encircle{\footnotesize 1}] $\textsf{TX}_i$ downloads $\textsf{RX}_j$ model parameters $[\vphi_j, \vgamma_j]$ while measuring $\varepsilon_{ij}$ under uplink-downlink channel reciprocity.

\item[\encircle{\footnotesize 2}] $\textsf{TX}_i$ partially freezes the downloaded model parameters, and locally fine-tunes a virtual transceiver $\textsf{TRX}_{ij}=[\vtheta_i, \textsf{Freeze}_\ell(\vphi_j), \vgamma_j]$ using $\mathcal{X}_i$ under an applying measured crossover probability $\varepsilon_{ij}$, yielding a fine-tuned virtual transceiver $\textsf{TRX}'_{ij}=[\vtheta_i', \vphi_j', \vgamma_j]$.

\item[\encircle{\footnotesize 3}] $\textsf{TX}_i$ uploads the fine-tuned unfrozen model parameters, i.e., non-zero elements of $[\vphi'_j] - [\textsf{Freeze}_\ell(\vphi_j)]$, to $\textsf{RX}_j$.

\item[\encircle{\footnotesize 4}] $\textsf{TX}_i$ transmits the SR $\vz_{ij}$ encoded using $\vtheta_i'$, and $\textsf{RX}_j$ decodes the received $\hat{\vz}_{ij}$ using $[\vphi_j',\vgamma_j]$.

\end{enumerate}

In \encircle{\footnotesize 2}, the function $\textsf{Freeze}_\ell(\cdot)$ freezes the $\ell$-th layers of $\vphi_j$ with $\ell\in\{0, 1, 2, \cdots, L\}$ counting from the last layer. This counting order yields less performance degradation based on our experiments. The case $\ell=0$ implies that $\vphi_j$ is entirely re-trained. In this case we additionally apply parameter re-initialization before re-training, which improves the performance based on our experiments.

The fine-tuning loss function of SLF follows from the standard VQ-VAE loss function $\mathcal{L}$ given as follows \cite{NIPS2017_7a98af17}:
\begin{align}
    \mathcal{L} \!=\! \|{{\vx}_i} - {{\hat{\vx}_{ij}}}\|_2^2 + \|\hat{\vz}_{ij} - sg[{\vz}'_{ij}]\|_2^2 
    + \lambda_c \|sg[\hat{\vz}_{ij}] - {\vz'}_{ij}\|_2^2 . \label{Eq:Loss}
\end{align}
The terms $\lambda_c$ is a constant hyper-parameter for the codebook commitment loss, and $sg[\cdot]$ is the stop-gradient operator for ensuring differentiability. 
Note that the first term in \eqref{Eq:Loss} coincides with the reconstruction task's MSE. For the classification task, the classifier $\vgamma_j$ is separately pre-trained and frozen, and no additional loss term is taken into account.
\vspace{-5pt}

\section{Numerical Evaluation}\label{numerical}
To validate the effectiveness of SLF, we consider three transceivers $\textsf{TRX}_{i}$ with $i\in\{1,2,3\}$, each of which consists of a pair of the jointly pre-trained transmitter $\textsf{TX}_i$ and $\textsf{RX}_i$. The $\textsf{TX}_i$ and $\textsf{RX}_i$ components comprise $3$ convolutional layers each, with Relu activation functions applied to all layers except the final one. The codebook for the VQ-VAE architecture was structured as a $16 \times 16$ embedding layer. As for the classifier responsible for the task-specific aspect, it consisted of a sequence of $2$ consecutive convolutional layers followed by $2$ linear layers. A common design pattern was employed by incorporating a maxpooling layer after each convolutional layer output. Hence, in the experimental setting, the freeze parameter $\ell$ can be configured within the range of $0$ to $4$, incorporating $3$ convolution layers and $1$ embedding layer.

The pre-training environment for $\textsf{TRX}_i$ is characterized by the training datasets and the crossover probability $\varepsilon_i$ of the DMCs, which are by default given as below.
\begin{itemize}
\item $\textsf{TRX}_{1}$: MNIST dataset with $\varepsilon_1 = 10^{-5}$
\item $\textsf{TRX}_{2}$: MNIST dataset with $\varepsilon_2 = 10^{-1}$
\item $\textsf{TRX}_{3}$: CIFAR-10 dataset with $\varepsilon_3 = 10^{-5}$
\end{itemize}
We consider the semantics misaligned transceivers $[\textsf{TX}_1,\textsf{RX}_2]$ and $[\textsf{TX}_1,\textsf{RX}_3]$ to study the impacts of channel and source data dissimilarities, respectively.

During the pre-training phase (before SLF), each dataset is divided into training and test datasets in the ratio of $8:2$. The batch size is $128$, and the optimizer is Adam with the learning rate $10^{-3}$. During the fine-tuning phase (in SLF) with $\ell\geq 1$, we follow the same setting, but the learning rate is reduced to $10^{-4}$. If $\ell=0$, the entire parameters are re-trained, in which we additionally apply re-initialization and use the original learning rate $10^{-3}$.

For a reconstruction task, $\textsf{TRX}_{ij}$ has a bipartite structure: $\textsf{TX}_i=[\vtheta_i]$ and $\textsf{RX}_j=[\vphi_j]$, where $\vtheta_i$ and $\vphi_j$ follow the encoder and the decoder of a VQ-VAE NN. For a classification task, $\textsf{TRX}_{ij}$ has a tripartite structure: $\textsf{TX}_i=[\vtheta_i]$ and $\textsf{RX}_j=[\vphi_j,\vgamma_j]$, where $\vgamma_j$ is a pre-trained classifier NN. In the codebook $\mathcal{C}_{ij}$, the number $K$ of codewords is set as $16$, which corresponds to the output and input dimensions of $\vtheta_i$ and $\vphi_j$, respectively.

\textbf{Impact of Channel Dissimilarity}.\quad 
With $[\textsf{TX}_1,\textsf{RX}_2]$, we assume that channel statistics, i.e., $\varepsilon_{12}$, are known at $\textsf{TX}_1$. Fig. \ref{Graph 3} shows that SLF reduces the baseline MSE from a minimum of $84.6\%$ to $92.1\%$. For classification, we increased the Top-1 accuracy by at least $67.1\%$ and up to $70.9\%$. From the results showing the performance of various cases of $\varepsilon_{12}$ and the freeze parameter $\ell$ for $\textsf{TX}_1$ and $\textsf{RX}_2$, we can observe the following. The closer the environment $\textsf{RX}_2$ was trained in and the new environment $\textsf{TRX}_{12}$ is facing, the more effectively the freeze parameter $\ell$ works. This means that we can further increase communication efficiency depending on the CSC problem, which we explain in more detail later when discussing latency.

\begin{table}[t]
    \centering
    \caption{Communication and Computation Costs of SLF with respect to the number $\ell$ of frozen layers.}
    \label{table: numerical result}
    \resizebox{\columnwidth}{!}{
    \begin{tabular}{cccccccc}    
        \toprule
        \multicolumn{2}{c} {} & $\ell=0$ & $\ell=1$ & $\ell=2$ & $\ell=3$ & $\ell=4$\\ \hline
        \multicolumn{2}{c}{\begin{tabular}[c]{@{}c@{}} {DL Payload} Size\\ $[\text{kBytes}]$ \end{tabular}} & 114 & 114 & 114 & 114 & 114\\
        \hline
        \multicolumn{2}{c}{\begin{tabular}[c]{@{}c@{}}  \textbf{DL} \textbf{Latency} $[\text{s}]$ \\ \encircle{\scriptsize 1}\end{tabular}} & \textbf{0.456} &  \textbf{0.456} & \textbf{0.456} & \textbf{0.456} & \textbf{0.456}\\
        \hline
        \multicolumn{2}{c}{\begin{tabular}[c]{@{}c@{}} {Fine-Tuning} Comput.\\ $[\text{TFLOPS}]$\end{tabular}} & 79.02 & 64.28 & 48.19 & 50.63 & {47.32}\\
        \hline
        \multicolumn{2}{c}{\begin{tabular}[c]{@{}c@{}} \textbf{Fine-Tuning} \textbf{Latency} $[\text{s}]$\\ for Reconstruction \encircle{\scriptsize 2} \end{tabular}} & 2.63 & 2.14 & 1.60 & 1.68 & \textbf{1.57}\\
        \hline
        \multicolumn{2}{c}{\begin{tabular}[c]{@{}c@{}} {Fine-Tuning} Comput.\\ $[\text{TFLOPS}]$\end{tabular}} & 75.43 & 67.49 & 39.69 & 37.31 & {34.17}\\
        \hline
        \multicolumn{2}{c}{\begin{tabular}[c]{@{}c@{}} \textbf{Fine-Tuning} \textbf{Latency} $[\text{s}]$\\ for Classification \encircle{\scriptsize 2}$'$ \end{tabular}} & 2.51 & 2.24 & 1.32 & 1.24 & \textbf{1.13}\\
        \hline
        \multicolumn{2}{c}{\begin{tabular}[c]{@{}c@{}} {UL Payload} Size\\ $[\text{kBytes}]$ \end{tabular}} & 114 & 76 & 22 & 2 & {0}\\
        \hline
        \multicolumn{2}{c}{\begin{tabular}[c]{@{}c@{}}  \textbf{UL} \textbf{Latency} $[\text{s}]$\\ \encircle{\scriptsize 3}\end{tabular}} & 0.456 & 0.304 & 0.088 & 0.008 &    \textbf{0}\\
        \bottomrule
        \bottomrule
        \multicolumn{2}{c}{\begin{tabular}[c]{@{}c@{}} \textbf{Recovery Time} $[\text{s}]$\\ \encircle{\scriptsize 1} + \encircle{\scriptsize 2} +  \encircle{\scriptsize 3}    \end{tabular}} & 3.542 & 2.900 & 2.144 & 2.144 & \textbf{2.026}\\
        \hline
        \multicolumn{2}{c}{\begin{tabular}[c]{@{}c@{}} {Reconstruction} \\\textbf{MSE}  \end{tabular}} & \textbf{0.087} & 0.103 & 0.113 & 0.121 & 0.123\\
        \hline
        \multicolumn{2}{c}{\begin{tabular}[c]{@{}c@{}} \textbf{Recovery Time} $[\text{s}]$\\ \encircle{\scriptsize 1} + \encircle{\scriptsize 2}$'$ + \encircle{\scriptsize 3} \end{tabular}} & 3.422 & 3.000 & 1.864 & 1.704 & \textbf{1.586}\\
        \hline
        \multicolumn{2}{c}{\begin{tabular}[c]{@{}c@{}} Classification\\ \textbf{Accuracy} $[\text{\%}]$\end{tabular}} & \textbf{98.55} & 98.04 & 97.83 & 97.01 & $96.90$\\
        \bottomrule
    \end{tabular}}
\end{table}

\begin{figure}[t!]
    \centering
        \includegraphics[width=0.85\columnwidth]{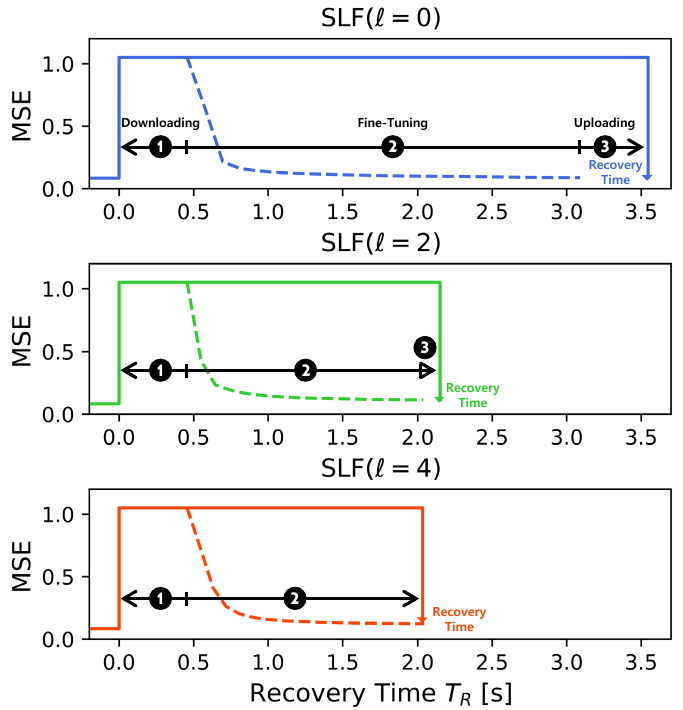}
        \label{recovery time and MSE}
        \vspace{-10pt}
     \caption{Recovery time versus reconstruction MSE corresponding to parameter $\ell$ in SLF.}
     \label{Graph 6}
     \vspace{-15pt}
\end{figure}

\textbf{Impact of Source Data Dissimilarity}.\quad 
The right side of each subfigure in Fig. \ref{Graph 3} shows the SLF results according to the CSC problems caused by the difference in the trained data source environment. For a meaningful analysis according to the data distribution, we assume that the dataset ${\mathcal{X}_1}'$ sent by $\textsf{TX}_1$ follows the below formula.
\begin{align}
    {\mathcal{X}_{1}}' = (1 - \lambda_{13})\mathcal{X}_{\text{1}} + \lambda_{13}\mathcal{X}_\text{3} ,
\end{align}
where $\lambda_{13}$ is defined as data dissimilarity. In this simulation, we use $\mathcal{X}_{\text{1}}$ as the MNIST dataset and $\mathcal{X}_{\text{3}}$ as the CIFAR$10$ dataset. To construct a classifier for the same comparison, we used the classifier $\vgamma_3$, which was trained on both MNIST and CIFAR$10$ datasets.

The results show when the dataset ${\mathcal{X}_1}'$ has low data dissimilarity with ${\mathcal{X}_3}$, $\textsf{RX}_3$ has trained, adjusting the freeze parameter $\ell$ according to the trade-off is abled. When $\lambda_{13} = 0$, the difference in SLF results based on the freeze parameter $\ell$ is $8.4\%$ in the reconstruction task and $91.8\%$ in the classification task. As the data dissimilarity becomes larger, the difference by the freeze parameter $\ell$ becomes smaller and finally converges to the same.

\textbf{Recovery Time}.\quad
We define recovery time as $T_R$ which takes to resolve a CSC problem, details as from the time taken from a point CSC problem occurred to $\textsf{RX}_j$ received the ${\vphi}_j'$ from $\textsf{TX}_i$ completely. $T_R$ varies depending on the uplink (UL) and downlink (DL) channel capacity between the $\textsf{TX}_i$ and $\textsf{RX}_j$, and the computing power of the $\textsf{TX}_j$. We set those values to $2$Mbps, $2$Mbps, and $30$TFLOPS, respectively. The simulation environment is when $\textsf{TX}_1$ and $\textsf{RX}_2$ apply SLF to resolve the CSC problem at $\varepsilon_{12} = 10^{-5}$. Tab. \ref{table: numerical result} shows the difference in recovery time according to $\textsf{RX}_2$'s task-specific $\vgamma_2$. SLF with freeze parameter $\ell = 4$ had the fastest $T_R$ of $2.026$s and $1.586$, respectively. But the reconstruction MSE and accuracy were the worst. This trade-off shows the potential for the proposed SLF to be dynamically adapted based on recovery time and performance.

\section{Conclusion}
In this paper, we addressed the issue of semantic misalignment arising from the nature of NNs in DeepJSCC with AI transceiver. To solve this problem, we proposed Split Learning with layer Freezing (SLF) and analyzed various scenarios depending on the key parameter in the SLF operation.
With this promising solution, our study highlights the significance of achieving interoperability when constructing a communication system using an AI transceiver in a multi-user communication environment.

\bibliographystyle{ieeetr}
\bibliography{main}
\end{document}